\documentclass{article}

\usepackage{arxiv}

\usepackage[utf8]{inputenc} 
\usepackage[T1]{fontenc}    
\usepackage{hyperref}       
\usepackage{url}            
\usepackage{booktabs}       
\usepackage{amsfonts}       
\usepackage{nicefrac}       
\usepackage{microtype}      
\usepackage{lipsum}		
\usepackage{graphicx}
\usepackage{natbib}
\usepackage{doi}
\usepackage{dirtytalk}
\usepackage[table,x11names]{xcolor}
\usepackage{amsmath}
\usepackage{booktabs}
\usepackage{adjustbox}
\usepackage{amsthm}
\usepackage{colortbl}
\usepackage{multirow}

\title{Latents of latents to delineate pixels: hybrid Matryoshka autoencoder-to-U-Net pairing for segmenting large medical images in GPU-poor and low-data regimes}

\author{
    {\hspace{1mm}Ariba Khan,}
    {\hspace{1mm}Sawera Hanif,}
    \href{https://orcid.org/0000-0003-0638-9689}{\includegraphics[scale=0.06]{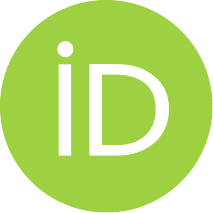}\hspace{1mm}Tahir Syed} \\
    School of Mathematics and Computer Science, Institute of Business Administration Karachi, Pakistan \\
    \texttt{\{tahirqsyed\}@gmail.com} }

\renewcommand{\shorttitle}{\textit{MatAE}  here goes short title}

\hypersetup{
pdftitle={A template for the arxiv style},
}

\begin{document}
\maketitle

\begin{abstract}

Medical images are often high-resolution and lose important detail if downsampled, making pixel-level methods such as semantic segmentation much less efficient if performed on a low-dimensional image. We propose a low-rank Matryoshka projection and a hybrid segmenting architecture that preserves important information while retaining sufficient pixel geometry for pixel-level tasks. We design the Matryoshka Autoencoder (MatAE-U-Net) which combines the hierarchical encoding of the Matryoshka Autoencoder with the spatial reconstruction capabilities of a U-Net decoder, leveraging multi-scale feature extraction and skip connections to enhance accuracy and generalisation. We apply it to the problem of segmenting the left ventricle (LV) in echocardiographic images using the Stanford EchoNet-D dataset, including 1,000 standardised video-mask pairs of cardiac ultrasound videos resized to 112x112 pixels. The MatAE-UNet model achieves a Mean IoU of 77.68\%, Mean Pixel Accuracy of 97.46\%, and Dice Coefficient of 86.91\%, outperforming the baseline U-Net, which attains a Mean IoU of 74.70\%, Mean Pixel Accuracy of 97.31\%, and Dice Coefficient of 85.20\%. The results highlight the potential of using the U-Net in the recursive Matroshka latent space for imaging problems with low-contrast such as echocardiographic analysis.

\end{abstract}


\section{Introduction}
This project focuses on developing U-Net models for Left Ventricular Segmentation, an important task in cardiovascular diagnoses. We utilized a Vanilla U-Net and a hybrid neural network architecture that combines the hierarchical encoding capabilities of the Matryoshka Autoencoder (MatAE) with the spatially precise reconstruction abilities of a U-Net Decoder. Designed for tasks requiring high-resolution outputs, such as medical image segmentation and image restoration, this approach leverages multi-scale feature extraction and skip connections to achieve superior performance in both accuracy and generalization.

\section{Experiments}

\textbf{Dataset.} The Stanford EchoNet-D dataset is a curated collection of cardiac ultrasound videos designed for segmentation and diagnostic analysis. The dataset comprises 1,000 video-mask pairs provided in a standardized format. Each video represents a sequence of echocardiographic frames and is paired with manually annotated segmentation masks for cardiac structures.

The videos were standardized to ensure uniform resolution and preprocessing, with each frame resized to an input size of 112 x 112 pixels. This resolution balances the need for computational efficiency with the preservation of clinically relevant details required for segmentation tasks.

The EchoNet-D dataset serves as an excellent benchmark for cardiac imaging research due to its high-quality annotations, temporal consistency across frames, and focus on real-world clinical scenarios. Its standardized format and resolution make it well-suited for deep learning applications, including architectures like the Vanilla U-Net and MatAE-U-Net.

\section{Methodology}

This section describes the methodology adopted to train a neural network for segmenting the left ventricle from echocardiographic images, utilizing a dataset of 1000 manually annotated images.

\textbf{Dataset Preparation}

The dataset consisted of echocardiographic images and their corresponding manually annotated segmentation masks:

Annotation Process

The authors, AK and SH,  annotated the masks using medical imaging software, marking the boundaries of the left ventricle. SH is a clinical expert and guided AK regarding the boundaries and helped develop an understanding of the walls and nature of the shape of the left ventricle for AK to annotate images. These masks were stored as binary images, with pixel values of 1 indicating the left ventricle and 0 representing the background.

\textbf{Neural Network Architecture}

Vanilla U-Net

The Vanilla U-Net is a symmetric encoder-decoder architecture designed for segmentation tasks, particularly in the field of medical imaging. Its simplicity and efficiency have established it as a benchmark model for comparison in image segmentation studies. The architecture is composed of an encoder, a bottleneck layer, and a decoder, all of which work together to extract features, capture high-level abstractions, and reconstruct detailed segmentation masks.

The encoder in the Vanilla U-Net consists of a series of convolutional blocks, each containing two convolutional layers with 3x3 filters. These layers are followed by ReLU activation functions, which enable the network to learn meaningful spatial features. At the end of each block, a max pooling layer with a 2x2 kernel is applied to downsample the spatial dimensions. With each subsequent block, the feature map depth doubles while the spatial resolution is halved, creating a hierarchical representation of features. This progressive downsampling allows the encoder to capture increasingly abstract and high-level patterns within the input data.

At the center of the U-Net lies the bottleneck layer, which represents the deepest level of abstraction in the architecture. This layer processes the encoded feature maps to distill high-level features that are critical for effective segmentation.

The decoder mirrors the encoder's structure and is responsible for reconstructing the segmentation mask. Upsampling is achieved using transpose convolutions, which restore the spatial resolution of the feature maps. Additionally, the Vanilla U-Net employs skip connections, where feature maps from the encoder are concatenated with corresponding layers in the decoder. This ensures that low-level spatial details lost during the downsampling process are preserved, which is crucial for accurately delineating boundaries in the segmentation mask. Each decoder block consists of two convolutional layers with 3x3 filters, and as the upsampling progresses, the feature map depth decreases while the spatial resolution increases.

The final output of the network is generated through a 1x1 convolutional layer, which creates a binary segmentation mask. This mask assigns a likelihood score to each pixel, indicating its membership in a specific class.

The Vanilla U-Net’s straightforward design makes it highly effective for datasets of moderate complexity. Its ability to preserve fine-grained spatial details through skip connections is a key strength, ensuring precise boundary detection. However, the architecture does have limitations. Its symmetric design may struggle with datasets requiring a deeper contextual understanding or more complex hierarchical feature representations.

Despite these challenges, the Vanilla U-Net remains a foundational architecture in the field of image segmentation. Its reliability and computational efficiency make it an ideal baseline model, paving the way for more advanced architectures like the MatAE-U-Net, which build upon its core principles to address its limitations.
\section{Matryoshka Autoencoder (MatAE) U-Net}

This hybrid convolutional neural network consists of:

Matryoshka Autoencoder (MatAE): Inspired by the Matryoshka dolls, where each smaller doll fits perfectly within the larger one. This concept inspired the encoding of the input data into progressively smaller feature representations to capture the hierarchy of features from global to fine-grained.
This is for the encoding process where the input is passed through a series of convolutional layers to reduce the dimensionality while extracting meaningful features\cite{kusupati2022matryoshka}.
Each layer uses MaxPooling to halve the spatial dimensions creating a feature hierarchy.
The outputs from each stage are stored as features maps to be later used by the decoder for the reconstruction.

This process ensures the preservation of outputs from every stage of encoding where MatAE ensures that the decoder has access to both low-level and high-level features which is critical for our segmentation task.

\textbf{U-Net:} This functions as the decoder of this hybrid neural network, where the features maps from MatAE serve as inputs to the U-Net decoder.
Skip Connections: Feature maps from the MatAE (encoder) correspond to the layers in the decoder. These connections help recover the spatial details lost during encoding.
Upsampling via Transpose Convolutions: They gradually restore the spatial resolution to match the input dimensions. Transpose convolutions are the process of sliding the input over the kernel unlike the convolutional operation where the kernel slides over the input. This reversal of operation results in upsampling where a smaller input produces a larger output through element-wise multiplication and concatenation.
Feature Refinement: Each upsampled layer is concatenated with its corresponding encoder feature map followed by convolution blocks for its refinement.
The output is reconstructed by a final convolution layer

\textbf{Behind-the-Scenes Logic}

The Matryoshka Autoencoder was inspired by the suggestion of the instructor, Dr. Tahir Syed. He suggested looking into Matryoshka Representation Learning (MRL). When we studied the concept behind it, we were deeply intrigued by the potential of incorporating the MRL in an autoencoder, MatAE, and then incorporating the MatAE as the encoder part of our U-Net (MatAE-UNet)\cite{ouyang2019echonet}.

This model leverages the contextual nesting of the MRL and the latent representation of the autoencoders creating a nested structure of embeddings. The decoder block received the input after passing through the convolutional block for the processing of concatenated features.

The way this works is that MRL learns the features at different spatial scales by leveraging the nested encoding concept incorporated into the MatAE, producing each feature map at a specific resolution based on the ‘Russian doll’-like architecture building a hierarchy, resulting in the production of feature maps containing increasingly finer information in each layer - contextually nested. These feature maps are progressively used in the U-Net decoder through skip connections to ensure both abstract (high level) and spatial (low level) features are provided to the decoder arm of the hybrid neural network.

The MatAE-UNet provides a hierarchical representation learning with multi-resolution encoding and reconstruction by efficiently reducing the dimensions of the input and enabling appropriate upsampling using transpose convolution and skip connections to improve the interpretability and robustness of the model. This model is also scalable and versatile in it’s functionality. The benefit of this end-to-end training process allows the MatAE to learn features that are useful for the decoder.

During the design of this model, we struggled with trying to incorporate the MRL concept with the autoencoders to create the MatAE and once we had achieved a functional MatAE, to incorporate it into the U-Net. We went back and forth between coding the encoder arm of the U-Net such that it worked seamlessly with the MatAE as the encoder.
\subsection{ Data Preprocessing}

The dataset is composed of echocardiographic videos which we converted into frames. Due to memory and computational constraints, we decided to store random frames from the videos. We decided to save one random frame per video due to the available memory running out in cases where we kept more frames.

We used two GPU T4 processors available on Kaggle for our project.
\section{Training Pipeline}

\textbf{Data Splitting:}
The trial run for the Vanilla U-Net was on 250 annotated images. The final dataset had 1014 frames that were annotated by the authors, split into 1000 train and 14 test images. This decision was made to provide our models as many images as we could for better training.

\textbf{Optimization:}
We used the Adam optimizer, learning rate and Binary Cross-Entropy Loss function for our models. The reason for using the Binary Cross-Entropy was used because we did not use the sigmoid function.

\textbf{Evaluation Metrics:}
Pixel Accuracy, Dice Coefficient and Intersection over Union (IoU) were used to assess the left ventricle segmentation accuracy of the models.

\textbf{Implementation Details}

The training was conducted using PyTorch, with custom Dataset and DataLoader classes for efficient data management. The training process involved multiple epochs, with periodic evaluations on the test set.

\subsection{Analysis}

\textbf{Overview}

The figures illustrate echocardiographic images of the left ventricle (LV) paired with their segmentation results. Each row includes the original echocardiographic image, a ground truth segmentation mask (manually annotated), the predicted segmentation mask generated by an automated model, and the final left ventricular segmentation. The goal of the segmentation process is to accurately delineate the left ventricle, a critical structure for assessing cardiac function and identifying pathological conditions.

\textbf{Ground Truth Masks}

The ground truth masks represent manually annotated segmentation of the LV and serve as the reference standard for evaluating the accuracy of the automated model. These masks demonstrate well-defined boundaries and precise delineation of the ventricular chamber.

From a clinical perspective, the ground truth masks provide essential anatomical insights. The normal morphology of the left ventricle, with its characteristic elongated shape and smooth borders, is evident in several examples. In contrast, certain masks suggest potential pathological conditions, such as ventricular dilatation or hypertrophy, based on deviations in size and contour, and the presence of emboli or ventricular wall pathologies can be detected in some masks. The annotations capture key structural features required for calculating clinically relevant parameters, including end-diastolic and end-systolic volumes, ejection fraction, and regional wall motion. The high-quality annotations ensure that both normal and pathological findings are well-represented for diagnostic and research purposes.

\textbf{Predicted Masks}

The predicted masks generated by the automated model exhibit a high degree of similarity to the ground truth masks, reflecting the model’s ability to identify the left ventricle's shape and boundaries in most cases. The predictions align well with the expected ventricular morphology, demonstrating the model’s capacity for reproducible and accurate segmentation in routine cases.

Despite the impressive results of the models, some discrepancies are observed:

Boundary Ambiguities: Predicted masks occasionally display less precise delineation along the ventricular borders, resulting in slightly irregular or jagged edges which is uncharacteristic of cardiac walls with the exception of certain pathological conditions such as traumatic injury - internal or external to the cardiac muscle, etc.

Under-Segmentation and Over-Segmentation: Certain masks either omit portions of the left ventricle or include extraneous regions, such as adjacent cardiac structures or papillary muscles. These inaccuracies, while minor, are clinically significant and may affect the calculation of clinical metrics like ventricular wall thickness and chamber volumes.

The model performs well in cases with clear boundaries, but it demonstrates limitations in handling more challenging scenarios, such as images with noise, shadowing, or anatomical variability.

\subsection{Comparison of Ground Truth and Predicted Masks}

The ground truth masks provide a gold standard for assessing the predicted masks. For normal ventricular anatomy, the predicted masks align closely with the ground truth, demonstrating the model’s robustness in identifying typical left ventricular contours. However, during the different phases of the cardiac cycle, particularly diastole when the heart is contracting or contracted, and in cases with pathological variations, such as hypertrophy or dilatation, the predicted masks show occasional deviations.


From a clinical standpoint, these discrepancies, particularly in boundary definition, may influence the accuracy of key echocardiographic measurements:

\textbf{End-Diastolic and End-Systolic Volumes:} Errors in segmentation may lead to underestimation or overestimation of ventricular volumes, impacting assessments of systolic function and ejection fraction.

\textbf{Regional Wall Motion Analysis:} Inconsistent boundary segmentation could affect the identification of wall motion abnormalities, a critical parameter for diagnosing ischemia or cardiomyopathy.
In spite of these limitations, the predicted masks demonstrate promising performance as a tool for automating segmentation, particularly in normal cases.

\textbf{Vanilla U-Net for Video Input:}

We experimented with Vanilla U-Net to test it on the echocardiographic video inputs which were .avi files, the standard input from the Stanford dataset. We took the masks annotated from the original videos and the generated masks from the Vanilla U-Net model. We then used those to generate videos of the segmented LV. 

For a model trained on only 1000 frames and having 1000 masks, the model performed significantly better than our expectations.
\begin{figure*}[!ht]
  \begin{minipage}{0.48\textwidth}
    \centering
    \includegraphics[width=\textwidth]{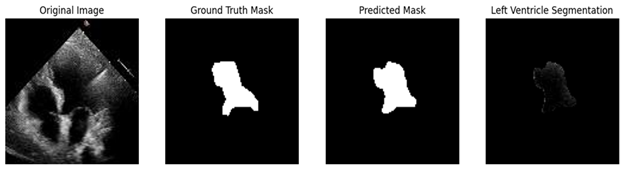}
    \caption{ Vanilla UNet Output}
    \label{fig:delta}
  \end{minipage}\hfill
  \begin{minipage}{0.48\textwidth}
    \centering
    \includegraphics[width=\textwidth]{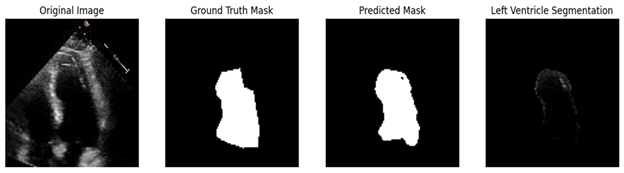}
    \caption{Vanilla UNet Output}
    \label{fig:guassian}
  \end{minipage}
  \begin{minipage}{0.48\textwidth}
    \centering
    \includegraphics[width=\textwidth]{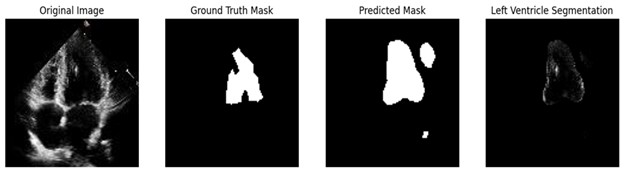}
    \caption{MatAE UNet Output}
    \label{fig:guassian}
  \end{minipage}
  \begin{minipage}{0.48\textwidth}
    \centering
    \includegraphics[width=\textwidth]{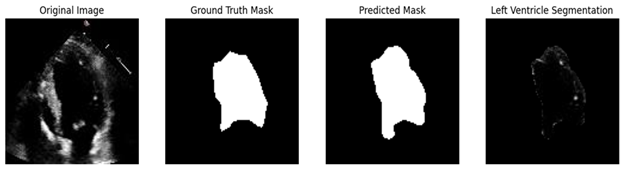}
    \caption{MatAE UNet Output}
    \label{fig:guassian}
  \end{minipage}
\end{figure*}

\begin{table}[!ht]
\centering
\caption{Vanilla U-Net for Video Input}
\arrayrulecolor{black}
\begin{tabular}{cc!{\color{black}}} 
\arrayrulecolor{black}\hline
\textbf{Metric}           & \textbf{Mean Average}  \\ 
\hline
Inference Time (s)        & 1.146                  \\ 
\arrayrulecolor{black}\hline
Average Memory Usage (MB) & 897.34                 \\ 
Peak Memory Usage (MB)    & 897.352                \\ 
Average CPU Usage (\%)    & 50.77                  \\ 
Peak CPU Usage (\%)       & 55.48                  \\ 
Average GPU Usage (\%)    & 1.37                   \\ 
Peak GPU Usage (\%)       & 1.37                   \\
\arrayrulecolor{black}\hline
\end{tabular}
\end{table}

\section{Results}
The Matryoshka UNet model achieved excellent segmentation performance, with a Mean IoU of 77.68\%, Mean Pixel Accuracy 97.46\%, and a Dice Coefficient of 86.91\% on the test set, demonstrating its effectiveness for medical image segmentation tasks.

The difference between the two models is wider with the Mean IoU and Mean Dice Coefficient, however, the Mean Pixel Accuracy are not too different. This emphasizes the balance between precision and recall indicating that Matryoshka UNet better captures the true regions of interest, and identifies the edges better than Vanilla UNet as indicated by the Mean Pixel Accuracy (Table 2, Table 3).

This methodology underscores the value of expert-annotated datasets and advanced neural network designs for precision in medical imaging applications.
\begin{table}[!ht]
\centering
\caption{Model Evaluation Metrics}
\begin{tabular}{cccc} 
\hline
\textbf{Model}  & \textbf{Mean IoU (\%)} & \textbf{Mean Dice Coefficient (\%)} & \textbf{Mean Pixel Accuracy (\%)}  \\ 
\hline
Vanilla UNet    & 74.70                  & 85.20                               & 97.31                              \\
Matryoshka UNet & 77.68                  & 86.91                               & 97.46                              \\
\hline
\end{tabular}
\end{table}

\begin{table}[!ht]
\centering
\caption{Model Comparison}
\begin{tabular}{lll} 
\hline
\textbf{Aspect}           & \textbf{Vanilla U-Net}                    & \textbf{MatAE-U-Net}                                   \\ 
\hline
\textbf{Encoder Design}   & Convolutional blocks with max pooling     & Hierarchical Matryoshka Autoencoder                  \\
\textbf{Skip Connections} & Between encoder and decoder at each stage & Between hierarchical encodings and decoder           \\
\textbf{Decoder Design}   & Symmetric to the encoder                  & Symmetric to encoder with added refinements          \\
\textbf{Strength}         & Simple, effective, fast to train          & Handles complex features, better generalization~     \\
\textbf{Use Case}         & Baseline segmentation model               & Advanced segmentation tasks requiring high accuracy  \\
\hline
\end{tabular}
\end{table}

\section{Conclusion}
In conclusion, the automated segmentation model provides a reliable approximation of left ventricular boundaries in most cases, aligning closely with manually annotated ground truth masks. With further optimization, the model holds significant potential to streamline echocardiographic analysis and assist clinicians in making accurate and timely diagnoses.

While MatAE-UNet exhibits strong potential for advanced segmentation tasks, the challenges outlined above highlight the need for careful design, tuning, and task-specific customization. Addressing these limitations is critical to fully realizing the capabilities of the MatAE-UNet architecture and ensuring its robustness in diverse application scenarios.

\bibliographystyle{unsrtnat}
\bibliography{MAE-arXiv} 

\begin{thebibliography}{2}
\providecommand{\natexlab}[1]{#1}
\providecommand{\url}[1]{\texttt{#1}}
\expandafter\ifx\csname urlstyle\endcsname\relax
  \providecommand{\doi}[1]{doi: #1}\else
  \providecommand{\doi}{doi: \begingroup \urlstyle{rm}\Url}\fi

\bibitem[Kusupati et~al.(2022)Kusupati, Bhatt, Rege, Wallingford, Sinha, Ramanujan, Howard-Snyder, Chen, Kakade, Jain, et~al.]{kusupati2022matryoshka}
Aditya Kusupati, Gantavya Bhatt, Aniket Rege, Matthew Wallingford, Aditya Sinha, Vivek Ramanujan, William Howard-Snyder, Kaifeng Chen, Sham Kakade, Prateek Jain, et~al.
\newblock Matryoshka representation learning.
\newblock \emph{Advances in Neural Information Processing Systems}, 35:\penalty0 30233--30249, 2022.

\bibitem[Ouyang et~al.(2019)Ouyang, He, Ghorbani, Lungren, Ashley, Liang, and Zou]{ouyang2019echonet}
David Ouyang, Bryan He, Amirata Ghorbani, Matt~P Lungren, Euan~A Ashley, David~H Liang, and James~Y Zou.
\newblock Echonet-dynamic: a large new cardiac motion video data resource for medical machine learning.
\newblock In \emph{NeurIPS ML4H Workshop}, pages 1--11, 2019.

\end{thebibliography}
\newpage
\appendix

\end{document}